\pdfoutput=1
\documentclass[
    preprint,                 
    superscriptaddress,       
    amsmath, amssymb,         
    aip,                      
    jcp,                      
    floatfix,                 
    letterpaper               
    superscript
]{revtex4-2}

\usepackage[english]{babel}              
\usepackage[T1]{fontenc}               
\usepackage{graphicx}                   
\usepackage[export]{adjustbox}          
\usepackage{caption}
\usepackage{subcaption}
\usepackage{dcolumn}                    
\usepackage{ragged2e}
\usepackage{bm}
\usepackage{braket}  
\usepackage{array} 
\usepackage{comment}                    
\usepackage{multirow}                   
\usepackage{placeins}                  
\usepackage{xcolor}                     
\usepackage[version=4]{mhchem}          
\PassOptionsToPackage{hyphens}{url}    
\usepackage[hyperindex,breaklinks]{hyperref} 
\usepackage{balance}
\usepackage{xcolor}
\usepackage{xspace}
\usepackage{natbib}
\usepackage{xr}

\newcommand{\Rtwo}{R\textsuperscript{2}}

\makeatletter
\newcommand{\ifpreprint}[1]{\@ifclasswith{revtex4-2}{preprint}{#1}{}}
\newcommand{\ifreprint}[1]{\@ifclasswith{revtex4-2}{reprint}{#1}{}}
\makeatother
\newcommand{\MYTITLE}{Computational Design of Low-Volatility Lubricants for Space Using Interpretable Machine Learning}

\begin{document}

\ifpreprint{
\begin{center}
    \LARGE\bfseries
    \MYTITLE
    
    \vspace{1em}
    
    \normalsize
    
    Daniel Miliate$^{1}$ and Ashlie Martini$^{1*}$  
    \vspace{1em}
    
    \small
    $^{1}$University of California Merced, Merced, CA 95343, United States \\
    $^{*}$amartini@ucmerced.edu
\end{center}
}
\ifreprint{
\title{\MYTITLE}

\author{Daniel Miliate}
    \affiliation{University of California Merced, Merced, CA 95343, United States}
\author{Ashlie Martini}
    \email{amartini@ucmerced.edu}
    \affiliation{University of California Merced, Merced, CA 95343, United States}
\maketitle
}


\begin{figure}[ht]
    \centering
    \includegraphics[width=3.125in]{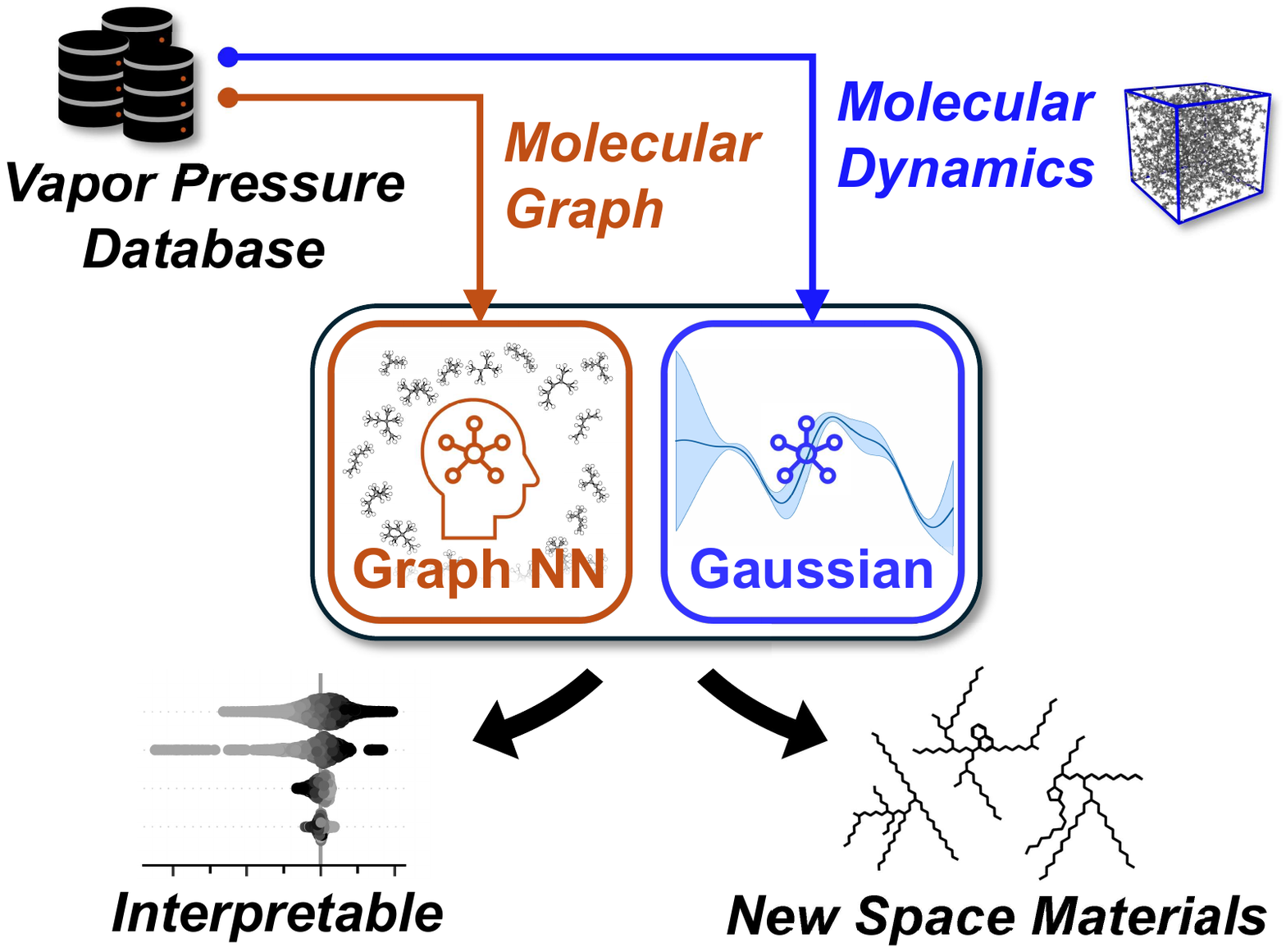}
    \label{fig:graphical_abstract}
\end{figure}

The function and lifetime of moving mechanical assemblies (MMAs) in space depend on the properties of lubricants. 
MMAs that experience high speeds or high cycles require liquid-based lubricants due to their ability to reflow to the point of contact. 
However, only a few liquid-based lubricants have vapor pressures low enough for the vacuum conditions of space, each of which has limitations that add constraints to MMA designs. 
This work introduces a data-driven machine learning (ML) approach to predicting vapor pressure, enabling virtual screening and discovery of new space-suitable liquid lubricants. 
The ML models are trained with data from both high-throughput molecular dynamics simulations and experimental databases. 
The models are designed to prioritize interpretability, enabling the relationships between chemical structure and vapor pressure to be identified. 
Based on these insights, several candidate molecules are proposed that may have promise for future space lubricant applications in MMAs. 
\ifpreprint{\newpage}
\section{\label{sec:Introduction}Introduction}
Moving mechanical assemblies (MMAs) in spacecraft, such as gimbals, actuators, and reaction wheels, require reliable lubrication to reduce friction and minimize wear~\cite{Fusaro1994-ak,Krishnan2015-vi,Jones2000-ti}.
On Earth, this requirement is commonly met using liquid-based lubricants like oils or greases.
In space, however, the performance of lubricants is constrained by exposure to vacuum , radiation, and wide operating temperatures.
Only a narrow subset of liquid-based lubricants can function in such conditions; common space lubricants include polyalphaolefins (PAOs), perfluoropolyethers (PFPEs), and multiply-alkylated cyclopentanes (MACs)~\cite{Fusaro1994-ak}.

Within this select family of lubricant chemistries, trade-offs must still be made to balance additive compatibility~\cite{Buttery2010-ge,Lu2009-qe,Clifford-Venier-Edward-W-Casserly-William-R-Jones-Jr-Mario-Marchetti-Mark-J-Jansen-Roamer-E-Predmore2002-bw,Carre1988-iq}, recommended operating temperatures~\cite{Marchetti2002-by}, and evaporative risk~\cite{Fusaro1992-rr}.
PFPEs are incompatible with many additive packages that improve wear resistance~\cite{Buttery2010-ge,Clifford-Venier-Edward-W-Casserly-William-R-Jones-Jr-Mario-Marchetti-Mark-J-Jansen-Roamer-E-Predmore2002-bw}, whereas MACs and PAOs can incorporate such additives~\cite{Lu2009-fv,Clifford-Venier-Edward-W-Casserly-William-R-Jones-Jr-Mario-Marchetti-Mark-J-Jansen-Roamer-E-Predmore2002-bw,Carre1988-iq}.
At extremely low temperatures, the viscosity increase exhibited by all three chemistries often require heaters, which add mass and power demands~\cite{Fusaro1994-ak,Scheidler2022-xn}.
PFPEs can operate at slightly lower temperatures than MACs and PAOs because the rate of change of viscosity with temperature is slower~\cite{Kent2023-bs,Jones2008-cm}, eliminating the need for heaters in some cases.
All liquid-based lubricants risk evaporation when in vacuum, but MACs and PFPEs are preferred over PAOs in vacuum exposed conditions because of their lower vapor pressures~\cite{Jones2006-xn}.
These tradeoffs constrain overall design flexibility, affecting mission cost and capability.
Expanding the pool of space-suitable liquid-based lubricants could alleviate these limitations.

Among the various properties of space lubricants, vapor pressure is one of the most critical to assess early in the qualifying processing, since even slow evaporation can lead to lubricant supply loss or contamination of sensitive components~\cite{Campbell1984-jn,ASTM-International2021-ba,European-Cooperation-for-Space-Standardization2008-ax}.
Methods for determining vapor pressure generally fall into three categories: direct experimental measurement with semi-empirical models, molecular modeling, and machine learning.

Experimental methods for measuring vapor pressure~\cite{Nathan-S-Jacobson-and-Benjamin-A-Kowalski2020-eq}, such as thermogravimetric analyzer~\cite{Nguyen2001-yr} or Knudsen effusion~\cite{Knudsen1909-oa, Oja1997-tr}, are based on evaporation in vacuum. 
Data measured using these instruments can be fit to semi-empirical models, like the Clausius-Clapeyron~\cite{Magomedov2023-lm, Brown1951-ab, Thompson2023-gd} and Antoine’s equations~\cite{Moshele2024-gj, Wisniak2001-bt}, to enable vapor pressure to be estimated using interpolation. 
However, the empirical constants are only valid for the materials and temperature ranges of the original measurements.
This limits their applicability to novel or untested molecules, leaving a gap in the ability to rapidly evaluate vapor pressures of new substances or expanded temperature ranges for heritage lubricants.

Molecular modeling has been used to estimate vapor pressure using either Monte Carlo (MC)~\cite{Factorovich2014-uk,Rane2013-db,Hens2018-np,Eggimann2020-bl} or Molecular Dynamics (MD)~\cite{Factorovich2014-uk,Muniz2021-et,Yakubovich2020-np,Quoika2024-nu} simulations.
In principle, these simulations model both vapor and liquid phase molecules in equilibrium, or vapor-liquid equilibrium (VLE). In VLE, the chemical potentials of the two phases are equal, and the vapor pressure can be determined by applying the virial theorem~\cite{Clausius1870-hv,Tsai1979-ej} to the molecules in the vapor phase~\cite{Eggimann2020-bl}. 
In addition to VLE, vapor pressure can be determined by simulating the pure vapor phase~\cite{Muniz2021-et}.
The vapor pressure of volatile liquids, either in VLE or pure vapor phase, can be performed using simulations with only a few thousands of atoms such that the simulations can be run with minimal computational costs and runtimes.
For larger, low-volatility molecules, however, phase transitions occur too infrequently to practically achieve VLE, often requiring specialized modeling strategies~\cite{Yakubovich2020-np}.
In addition to low-volatility limitation, the accuracy of MD simulations depends heavily on the accuracy of the force field~\cite{Quoika2024-nu}.
For novel molecular structures, especially those outside the chemical space for which force fields have been validated, simulation predictions can be unreliable.
These challenges in molecular simulations create a need for alternative approaches that can rapidly and accurately estimate vapor pressure without extensive parameterization.

Some of the limitations of experiments and molecular simulations are overcome with machine learning (ML) models.
Early ML models of vapor pressure used architectures such as multiple linear regression and simple neural networks~\cite{Basak1997-sy,Liang1998-ya,McClelland2000-hu,Katritzky1998-pw,Goll1999-kb,Basak2001-xr}.
These ML models relied on 2D molecular descriptors, or features, calculated from quantum chemistry. 
However, 2D molecular features cannot necessarily capture intermolecular interactions. 
Features derived from MD simulations have been used in Gaussian Process Regression (GPR) models with success, but have not been applied to vapor pressure~\cite{Panwar2024-uo}. 
In recent years, graph neural networks (GNN) have been used to predict vapor pressure~\cite{Santana2024-el,Lin2024-yo} and related properties like boiling point~\cite{Hoffmann2025-qe,Qu2022-my, Wang2023-dw} from graph representations of molecules.
In other applications, GNNs have offered an avenue to explain which structural factors contributed to the predicted property by determining the attribution of substructures~\cite{Wu2023-va}. 
However, current vapor pressure GNNs have not implemented these substructure masking techniques which inhibits interpretation of structure-property relationships.
Additionally, most existing ML models have been trained on datasets dominated by relatively volatile compounds, making them less useful for space-tribology applications where volatile compounds are rarely used, with the exception of highly specific use cases (e.g., gas bearings in Stirling converters~\cite{Wong2015-ye}). 
These factors highlight a need for ML models that are both trained from space-lubricant-relevant datasets and capable of explaining their predictions.

In this work, interpretable ML models for space lubricant discovery were developed using GPR and GNN architectures tailored for predicting low vapor pressure.
The GPR model was trained on a curated dataset combining experimental vapor pressure values with both static molecular descriptors (e.g., molecular weight) and dynamic molecular descriptors (e.g., radius of gyration) from short MD simulations. 
These descriptors provided single-point representations of each molecule and were incorporated as features in the GPR models.
In contrast, the GNN models leveraged molecular graph representations, where atoms served as nodes and chemical bonds as edges.
The graph representations were processed through a relational graph convolutional network, allowing the model not only to learn structural patterns (e.g., aromatic substructures or functional groups), and attribution methods can subsequently be applied to analyze the contributions of these substructures. 
Screening a large set of virtually generated molecular structures with this framework yielded several novel structures with predicted vapor pressures comparable to those of established space lubricants.
Altogether, this work demonstrates the value of interpretable ML in advancing space lubricant discovery while establishing transferable design principles that can guide the development of future space-qualified materials.
\section{Methods}
\subsection{Obtaining Target Data}
Molecular data were obtained from the NIST Chemistry WebBook~\cite{Linstrom1997-pf}, which included InChIKeys and Antoine's equation parameters.
In addition to the NIST data, Antoine's equation parameters for MACs were determined from reported vapor pressure measurements~\cite{UnknownUnknown-lo, Jones2004-zw, Venier2003-zy, Clifford-Venier-Edward-W-Casserly-William-R-Jones-Jr-Mario-Marchetti-Mark-J-Jansen-Roamer-E-Predmore2002-bw,Nguyen2001-no,UnknownUnknown-lj} and described in the Supplementary Information (SI). 

To ensure relevance to the range of chemistries in space-qualified lubricants, the dataset was restricted to molecules composed of carbon, hydrogen, oxygen, and fluorine. 
These elements were combined in patterns representative of PFPE-, MAC-, and PAO-like chemistries (H/C, H/C/O, H/C/F, C/F, and H/C/O/F).
A minimum of six carbon atoms per molecule was required to exclude small, highly volatile compounds unlikely to meet operational constraints for space lubrication.
Molecules differing only by geometric isomerism were consolidated by retaining the more energetically stable form.
This calculation was performed through geometry optimization of each isomer at the B3LYP~\cite{Becke1993-vx,Lee1988-fu,Becke1988-ko}/6-31G*~\cite{Hehre1972-lp} level of theory using Gaussian 16 Rev. B.01~\cite{Frisch2016-nf}.
Deuterated species were removed.
After these steps, 460 unique molecules from NIST remained. 
The number of molecules having H/C, H/C/O, H/C/F, C/F, and H/C/O/F chemistries were 188, 255, 11, 4, and 2, respectively. The distribution of molecular weights is visualized in Figure~\ref{fig:Molecular_Weight_Distribution}, where contributions from each chemistry are stacked and color-coded.

\begin{figure}[ht]
    \centering
    \includegraphics[width=3.125in]{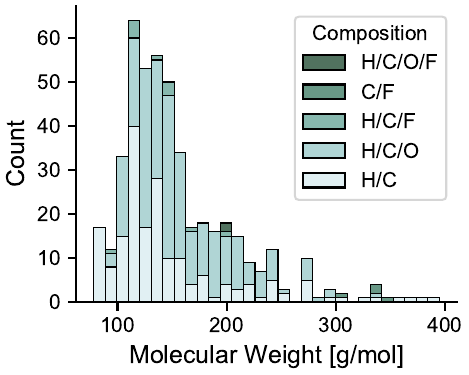}
    \caption{\justifying Stacked distribution of molecular weights for H/C, H/C/O, H/C/F, C/F, and H/C/O/F elemental compositions.}
    \label{fig:Molecular_Weight_Distribution}
\end{figure}
\FloatBarrier

The target data, vapor pressure, was calculated from Antoine's equation parameters from NIST and for MACs. 
The validated temperature ranges were not standardized, which meant that a single temperature could not be applied to all compounds. Instead, two datasets and featurization approaches were adopted.

In the first approach, vapor pressures were calculated at 20 evenly spaced temperatures between the minimum and maximum temperatures in the validated range.
The sampled temperatures were separated by at least 2 K to limit correlation between instances and remain within the practical temperature control limits of the MD simulations.
This resulted in a variable-temperature dataset containing 462 unique molecular structures and a total of 9,240 vapor pressure instances.  
In the second approach, a fixed-temperature dataset was created. Vapor pressure was calculated for 337 molecules at 387 K, the temperature most frequently represented in the available validated ranges. 

In both approaches, vapor pressure was $log_{10}$-transformed to balance loss contributions across many orders of magnitudes. 
Featurization of the two target datasets differed, being tailored to either the GPR or GNN model architectures. 

\subsection{Gaussian Process Regression Models}
Featurization for GPR models was designed to capture both static and dynamic molecular behavior, utilizing the variable-temperature dataset. 
Static descriptors, such as molecular weight and elemental counts, were calculated from SMILES strings~\cite{Weininger1988-ol} using RDKit~\cite{Landrum2024-mo}.
Dynamic descriptors, such as radius of gyration, were obtained from short non-reactive MD simulations using Py3dMD~\cite{Panwar2023-cv}. 
The MD workflow and simulation steps are detailed in the SI. 
The temperature used in the simulation corresponded to the temperature of the vapor pressure calculation in the dataset. 
The dynamic descriptors from the MD simulations vary at each sampled timestep. 
So, the PyL3dMD package was modified to output aggregate statistics: the average, standard deviation, minimum, 25th percentile, 50th percentile, 75th percentile, and maximum feature value for all molecules across all sampled frames.
This process produced more than 14,000 features, derived from dynamic and static molecular descriptors.

The large number of features required dimensionality reduction to improve the computational efficiency of model development. 
Three key techniques were applied. 
First, any feature with the same value for every molecule was removed. Second, redundant features were filtered by examining pairwise correlations; when two features had an \Rtwo~greater than 0.5, one was discarded. 
These two filters reduced the dataset to 35 static and 426 dynamic features. 
The third technique was performed using the Least Absolute Shrinkage and Selection Operator (LASSO)~\cite{Jiang2020-uf,Panwar2024-uo,Tibshirani1996-ck}. 
LASSO was chosen because it reduces model complexity by shrinking the coefficients of less influential features toward zero, effectively retaining only the most predictive variables.

To ensure robust evaluation of the remaining features with LASSO, the dataset was split into four partitions. 
The variable-temperature dataset contained multiple instances of the same molecular structure. 
However, molecules were kept to a single partition to avoid data leakage, i.e., not spread across multiple partitions. 
Stratification was applied to maintain a balanced distribution of target values across partitions. 
Because multiple vapor pressure values were available per molecule, the median was used for stratification. 
Three partitions were used for outer three-fold cross-validation (CV), and one was reserved as a holdout test set. 

Three LASSO models were trained using the three outer CV fold partitions. 
In each LASSO model, the two training folds were combined, and an inner 10-fold CV with random splits was implemented. 
The regularization parameter, $\lambda$, can be varied to decrease the coefficient of less predictive features. 
The final $\lambda$ value was determined using the one-standard-error (1SE) rule, balancing model accuracy and simplicity. 
This rule selected the $\lambda$ value corresponding with the mean squared error (MSE) of the 10-fold CV within one standard deviation of the minimum MSE observed~\cite{Panwar2024-uo}.
Figure~\ref{fig:LASSO_A} and~\ref{fig:LASSO_B} illustrate the $\lambda$ optimization curve and corresponding coefficient traces.
Features with non-zero coefficients in the three 1SE models were combined, resulting in a final set of 51 unique features (8 static and 43 dynamic).

\begin{figure}[ht]
     \centering
     \begin{subfigure}[b]{3.125in}
         \centering
         \includegraphics[width=\textwidth]{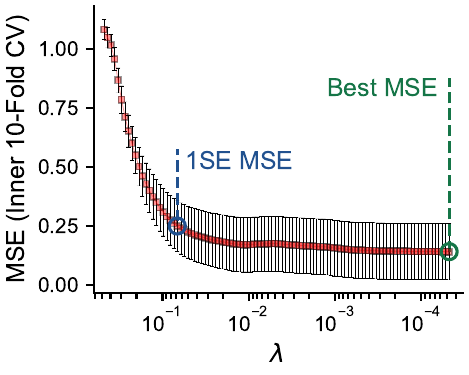}
         \caption{}
         \label{fig:LASSO_A}
     \end{subfigure}
     \hfill
     \begin{subfigure}[b]{3.125in}
         \centering
         \includegraphics[width=\textwidth]{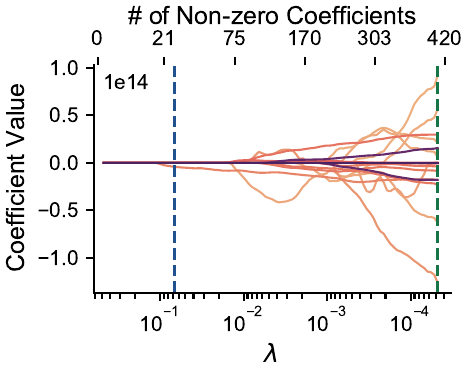}
         \caption{}
         \label{fig:LASSO_B}
     \end{subfigure}
     \hfill
     \caption{\justifying Representative (a) MSE vs $\lambda$ hyperparameter and (b) coefficient vs $\lambda$ hyperparameter from LASSO. In (a), the error bars are calculated from the inner 10-fold CV. The green dashed vertical line indicates the $\lambda$ hyperparameter value that resulted in the best (minimum) mean MSE. The dashed blue vertical line is one standard error (1SE) greater than the best MSE. In (b), feature coefficients are shown as a function of the $\lambda$ hyperparameter, with the corresponding number of non-zero coefficients on the top axis.} 
\end{figure}
\FloatBarrier

GPR models were trained with the variable-temperature dataset using systematically constructed feature subsets~\cite{Panwar2024-uo}. 
From the 51 available features, all unique one-, two-, three-, and four-feature combinations were evaluated, yielding 51, 1,275, 20,825 and 249,900 GPR models, respectively. 

The initial training of feature-subset GPR models was performed with three-fold cross-validation to determine the best feature subsets, using \textit{constant} and \textit{matern52} and the basis and kernel functions. Data was standardized for each feature subset using the training set for each cross-validation. 
All models with an average \Rtwo~> 0.7 were retrained and cross validated with all combinations of the basis functions, including \textit{none}, \textit{linear}, and \textit{pureQuadratic}, and kernel functions, including \textit{exponential}, \textit{squaredexponential}, \textit{matern32}, \textit{matern52}, \textit{rationalquadratic}, \textit{ardexponential}, \textit{ardsquaredexponential}, \textit{ardmatern32}, \textit{ardmatern52}, and \textit{ardrationalquadratic}. 
The best basis and kernel function combination was recorded for each feature set. 
Models with average \Rtwo~> 0.85 were retrained for the final evaluation. 
The three CV folds were used as the training set and benchmarked against the holdout test set. 
Models with a final evaluation \Rtwo~> 0.9 were kept and retrained using all the data available. 
Although models were filtered using \Rtwo, additional metrics are listed for completeness, such as root mean squared error (RMSE), mean absolute error (MAE), and mean absolute percent error (MAPE). Both the LASSO and GPR models were trained using MATLAB~\cite{Inc-The-Mathworks2024-hm}.

\subsection{Graph Neural Network Models}
The graph neural network (GNN) architecture and featurization used in this study followed one used to successfully predict aqueous solubility, mutagenicity, hERG-related cardiotoxicity and blood-brain barrier permeation~\cite{Wu2023-va}. 
The molecules from the fixed-temperature datasets were represented as molecular graphs, as opposed to static and dynamic molecular descriptors. 
Further discussion can be found in the original work~\cite{Wu2023-va}, but the GNN architecture is a relational graph convolution network (RCGN). 
The RCGN used the molecular graph representation as feature inputs with masking, i.e., graph nodes as atoms, functional groups, BRICS~\cite{Degen2008-bh} or Murcko substructures~\cite{Hu2016-vr,Bemis1996-py}. 
The masking allowed for the determination of attribution to the prediction for atoms and substructures. 

To train the RCGN, hyperparameters were explored first, and are available in Table~\ref{tab:Hyperparameters}. 
The fixed-temperature dataset was limited in size and instead used an 80:10:10 split for training, validation and holdout test splits.  
The best set of hyperparameters was then used to train 10 RCGN models with different weighting seeds. 

\setlength{\tabcolsep}{10pt}
\begin{table}[htbp]
  \centering
    \begin{tabular}{|c|>{}p{4cm}|}  
        \hline
        RCGN Hidden Layers & [[32, 32], [64, 64], [128, 128], [256, 256], [32, 32, 32], [64, 64, 64], [128, 128, 128], [256, 256, 256]] \\
        \hline
        FFN Hidden Features & [32, 64, 128, 256] \\ \hline
        FFN Drop Out & [0, 0.1, 0.2, 0.3, 0.4, 0.5, 0.6, 0.7, 0.8] \\ \hline
        RGCN Drop Out & [0, 0.1, 0.2, 0.3, 0.4, 0.5, 0.6, 0.7, 0.8] \\ \hline
        Learning Rate & [0.05, 0.01, 0.003, 0.001, 0.0003, 0.0001, 0.00001] \\ 
        \hline
    \end{tabular}
    \caption{\justifying Hyperparameter optimization for RCGN model.}
    \label{tab:Hyperparameters}
\end{table}
\section{\label{sec:Results_and_Discussion}Results and Discussion}
\subsection{GPR Model Interpretation}
GPR models with only one and two features exhibited poor predictive performance. 
Introducing a third feature markedly improved the model accuracy, yielding one model with an \Rtwo~> 0.9 on the holdout test set. Of the four-feature models, 34 had an \Rtwo~> 0.9 for the holdout test set. 
Details of the top-performing three- and four-feature models, including their features and evaluation metrics, are summarized in Table~\ref{tab:Best_GPR_Models}. A full table, including metrics for all 34 GPR models, is available in the SI. 

From the 34 four-feature models, the ten best models were selected to form an ensemble GPR model, designed to capture complementary predictive strengths and reduce individual model bias. 
The average prediction and associated uncertainties of the final evaluation are in Figure~\ref{fig:GPR_Parity}. 
The ensemble model demonstrated strong predictive performance, i.e., better than any individual model, with \Rtwo~of 0.946, RMSE of 0.296, MAE of 0.201, and MAPE of 6.44\% for the holdout test set. 
In addition to overall predictive performance, it is important that the GPR ensemble model predicts very low vapor pressure instances accurately. 
The 1,3-bis(2-octyldodecyl) cyclopentane molecule, which has the lowest vapor pressure instances in the test set, was predicted with \Rtwo~of 0.939, RMSE of 0.646, MAE of 0.573, and MAPE of 27.9\%.
The \Rtwo, RMSE and MAE signify that the model is still within one order of magnitude accuracy in predicting vapor pressure of the 1,3-bis(2-octyldodecyl) cyclopentane molecule. Following the final evaluation, all GPR models in the ensemble were retrained using the complete dataset to improve robustness and to examine relationships between molecular features and predicted vapor pressure.

\setlength{\tabcolsep}{10pt}
\begin{table*}[!htbp]
  \centering
    \resizebox{\textwidth}{!}{%
        \begin{tabular}{|c|c c c c|c c c c|}
            \hline
            \multirow{2}{*}{Features Names} 
                & \multicolumn{4}{c|}{Training Split} 
                & \multicolumn{4}{c|}{Test Split} \\ \cline{2-9}
                
            & \Rtwo & RMSE & MAE & MAPE (\%) 
              & \Rtwo & RMSE & MAE & MAPE (\%) \\ \hline
            
            $Temperature$ & \multirow{3}{*}{0.980} & \multirow{3}{*}{0.162} & \multirow{3}{*}{0.109} & \multirow{3}{*}{3.39} & \multirow{3}{*}{0.907} & \multirow{3}{*}{0.389} & \multirow{3}{*}{0.249} & \multirow{3}{*}{8.37} \\ 
            ${Density}_{mean}$ & & & & & & & & \\ 
            ${getawayHATSIP2}_{min}$ & & & & & & & & \\ 
            \hline

            $Temperature$  & \multirow{4}{*}{0.999} & \multirow{4}{*}{0.001} & \multirow{4}{*}{0.0004} & \multirow{4}{*}{0.014} & \multirow{4}{*}{0.940} & \multirow{4}{*}{0.312} & \multirow{4}{*}{0.231} & \multirow{4}{*}{7.41} \\ 
            ${Density}_{mean}$ & & & & & & & & \\ 
            ${getawayHATSIP2}_{min}$ & & & & & & & & \\ 
            ${FPSA3}_{mean}$ & & & & & & & & \\ 
            \hline

            $Temperature$ & \multirow{4}{*}{0.999} & \multirow{4}{*}{0.001} & \multirow{4}{*}{0.001} & \multirow{4}{*}{0.017} & \multirow{4}{*}{0.930} & \multirow{4}{*}{0.338} & \multirow{4}{*}{0.236} & \multirow{4}{*}{8.59} \\ 
            ${Density}_{mean}$ & & & & & & & & \\ 
            ${getawayHATSIP2}_{min}$ & & & & & & & & \\ 
            ${MoRSEu23}_{max}$ & & & & & & & & \\ 
            \hline

            $Temperature$ & \multirow{4}{*}{0.998} & \multirow{4}{*}{0.056} & \multirow{4}{*}{0.038} & \multirow{4}{*}{1.26} & \multirow{4}{*}{0.925} & \multirow{4}{*}{0.348} & \multirow{4}{*}{0.239} & \multirow{4}{*}{8.47} \\ 
            ${Density}_{mean}$ & & & & & & & & \\ 
            ${getawayHATSIP2}_{min}$ & & & & & & & & \\ 
            ${Density}_{std}$ & & & & & & & & \\ 
            \hline
        \end{tabular}
    }
    \caption{\justifying Top three- and four-feature subset GPR models performance metrics in the final evaluation. Only one three-feature model had an \Rtwo~greater than 0.9. The top four-features models contained the same three features, indicating high predictive ability.}
    \label{tab:Best_GPR_Models}
\end{table*}
\FloatBarrier

\begin{figure}
    \centering
    \includegraphics[width=3.125in]{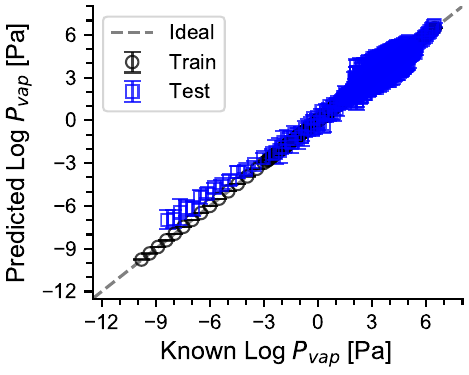}
    \caption{\justifying Parity plot of log vapor pressure ($P_{vap}$) of the temperature-dependent dataset containing 20 instances per molecule. The symbols and error bars represent the mean and standard deviation as predicted by the ten GPR models in the final evaluation.}
    \label{fig:GPR_Parity}
\end{figure}
\FloatBarrier

A histogram of the features is shown in Figure~\ref{fig:Feature_Histogram}. The ten of the nine four-feature models shared the same three features: $Temperature$, ${getawayHATSIP2}_{min}$, and ${Density}_{mean}$. The best-performing four-feature model, incorporating Temperature, ${getawayHATSIP2}_{min}$, ${Density}_{mean}$, and ${FPSA3}_{mean}$, was selected as a representative model for interpretability analysis.

\begin{figure}[!htbp]
    \centering
    \includegraphics[width=3.125in]{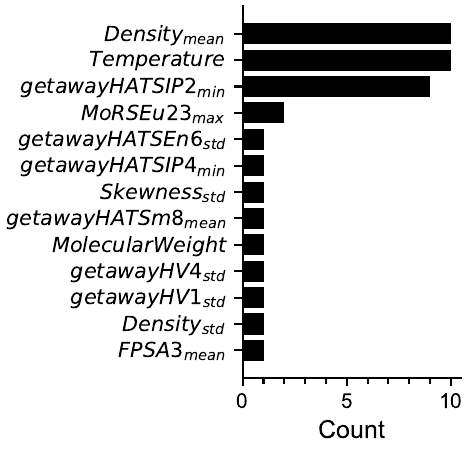}
    \caption{\justifying Histogram of all features in the top ten four-feature GPR models. $Temperature$, ${getawayHATSIP2}_{min}$, and ${Density}_{mean}$ appear in nearly all GPR models. }
    \label{fig:Feature_Histogram}
\end{figure}
\FloatBarrier

To understand how each feature influences vapor pressure predictions, SHapley Additive exPlanations (SHAP)~\cite{Wang2024-ji,Gawde2024-cf,Basu2022-xd} values were computed and visualized in Figure~\ref{fig:SHAP_Analysis_swarm}. 
The mean absolute SHAP values and corresponding SHAP swarm plot quantify each feature’s contribution to the model output.
According to the mean absolute SHAP values, the features contribute in the following order of importance:
$Temperature$ > ${getawayHATSIP2}_{min}$ > ${Density}_{mean}$ > ${FPSA3}_{mean}$.

\begin{figure}[!ht]
     \centering
     \begin{subfigure}[b]{3.125in}
         \centering
         \includegraphics[width=\textwidth]{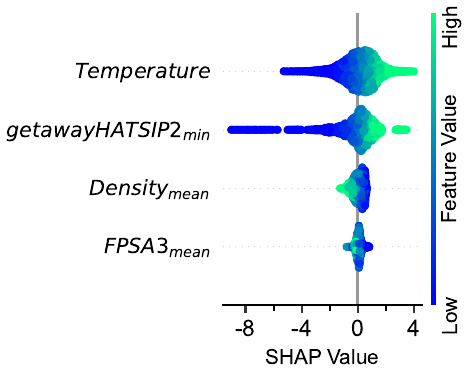}
         \caption{}
         \label{fig:SHAP_Analysis_swarm}
     \end{subfigure}
     \hfill
     \begin{subfigure}[b]{3.125in}
         \centering
         \includegraphics[width=\textwidth]{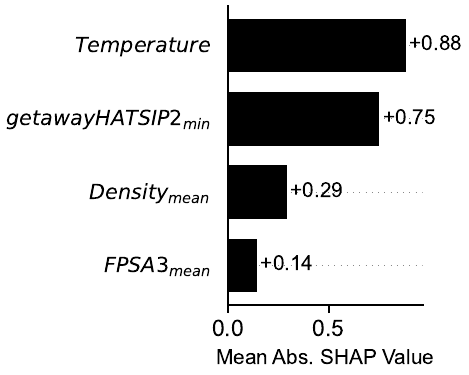}
         \caption{}
         \label{fig:SHAP_Analysis_MAS}
     \end{subfigure}
     \hfill
     \caption{\justifying (a) Representative swarm of SHAP values and (b) mean absolute SHAP values. Each row corresponds to a feature of the model. In (a), color indicates the feature value of the instance, with green colors indicating higher feature values and blue colors indicating lower ones. Negative SHAP values correspond to a decrease in the vapor pressure prediction. In (b), the mean of absolute SHAP values indicate the overall influence of the feature. In both (a) and (b), the rows are descending in order of importance, where the most important features are listed first.} 
\end{figure}
\FloatBarrier

In the swarm plot in Figure~\ref{fig:SHAP_Analysis_swarm}, negative and positive SHAP values correspond lower and higher predicted vapor pressures, respectively. 
The colors represent the feature values in relative terms. 
Simple features such as $Temperature$ and ${Density}_{mean}$ follow physical intuition: lower temperatures and higher densities correspond to lower vapor pressures. 
In contrast, the ${getawayHATSIP2}_{min}$ and ${FPSA3}_{mean}$ features are less straightforward to interpret.
The ${getawayHATSIP2}_{min}$ feature is derived from HATS autocorrelation~\cite{Consonni2002-py,Consonni2002-ff} with a lag of two and weighted by ionization potential. The ${FPSA3}_{mean}$ feature is computed from a molecule’s solvent-accessible surface area (SASA) for atoms having positive partial charge divided by total SASA~\cite{Stanton1990-gn,Stanton1992-yu}. 

From Figure~\ref{fig:SHAP_Analysis_swarm}, ${FPSA3}_{mean}$ has a non-monotonic relationship with predicted vapor pressure. 
Low ${FPSA3}_{mean}$ values correspond to higher vapor pressures, and high ${FPSA3}_{mean}$ values cluster near zero SHAP, indicating a neutral influence on predicting vapor pressure. 
However, intermediate ${FPSA3}_{mean}$ values correspond to lower vapor pressures. 
Chemically, high ${FPSA3}_{mean}$ values are typical of pure hydrocarbons, where numerous \(\delta^{+}\) hydrogens contribute to positively charged SASA. Low ${FPSA3}_{mean}$ values arise from fluorinated hydrocarbons, where most or all hydrogen atoms are replaced by \(\delta^{-}\) fluorine. Intermediate values come from the hydrocarbon molecules with hydroxyl functional groups, where the negative partial charge originates from \(\delta^{-}\) oxygen atoms. The chemical understanding of the ${FPSA3}_{mean}$ feature connects molecular geometry and surface polarity to vapor pressure through intermolecular cohesion. 
The ${getawayHATSIP2}_{min}$ feature also considers electron behavior, but instead the ionization potential weighting scheme instead of surface polarity. 

In general, interpretation of HATS autocorrelation features are based on combining leverage distances and weighting schemes. 
For leverage distances, atoms located farther from the geometric center contribute increase the value of the feature than those close to the geometric center. Here, the geometric center is defined by 3D atomic coordinates irrespective of mass.
The leverages are then weighted: for a weighting scheme utilizing atomic ionization potentials (F > O > H > C), F atoms far from the center will increase the value significantly more than any other atom. 
Additionally, because the HATS autocorrelation is summation based, more atoms are expected to further increase the feature value. 
This means that the ${getawayHATSIP2}_{min}$ feature is sensitive to both molecular size, shape, and weighting scheme. 
In our dataset, it is expected that fluorinated hydrocarbons would result in the highest feature value. 
However, more interpretation approaches for HATS features still need to be developed~\cite{Zapadka2022-rh}. 

\subsection{GNN Model Interpretation}
The parity plot of the GNN consensus model (of ten randomly seeded RCGN models) is shown in Figure~\ref{fig:GNN_Parity}. The GNN consensus model is able to predict the bulk of the data. For the few low vapor pressure molecules in the dataset, there was a larger variance in the prediction, but the mean prediction remains close to the known value. The \Rtwo, RMSE, MAE, and MAPE were 0.95, 0.23, 0.16, and 4.8\% for the training set, 0.91, 0.25, 0.20, and 5.7\% for the validation set, and 0.96, 0.31, 0.22, and 9.7\% for the test set, respectively. 

\begin{figure}[!htbp]
    \centering
    \includegraphics[width=3.125in]{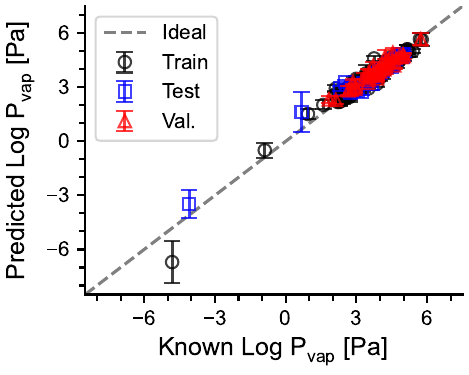}
    \caption{\justifying Parity plot of log vapor pressure ($P_{vap}$). The symbols and error bars represent the mean and standard deviation as predicted by the ensemble of ten RCGN models. Each marker is a unique molecule from the fixed-temperature dataset, where vapor pressure was calculated at 387 K.}
    \label{fig:GNN_Parity}
\end{figure}
\FloatBarrier

The substructure masking reveals the key connections between molecular structure and vapor pressure. The attribution of functional groups is shown in Figure~\ref{fig:fg_attribution}. Generally, when functional groups were present on the molecule, they contributed to lower vapor pressures. In particular, carboxylic acid, methyl ester, and methyl ketone functional groups were attributed with decreasing vapor pressure the most.
Fluorine-containing functional groups contributed to increasing vapor pressure or having almost no effect on the prediction. 
The conclusions from the GNN attributions are in agreement with the GPR analysis. 
Although this explanation is accurate for our dataset, it is important to note that PFPEs are absent from both the fixed-temperature and variable-temperature datasets. 
PFPEs, which are known to have extremely low vapor pressures and contain fluorine, are usually blends of chemical structures that vary greatly in molecular weight. 
Obtaining data for high-purity PFPEs would allow both models to better capture a wholistic behavior of compounds containing fluorine.

\begin{figure}[!htbp]
     \centering
     \includegraphics[width=3.125in]{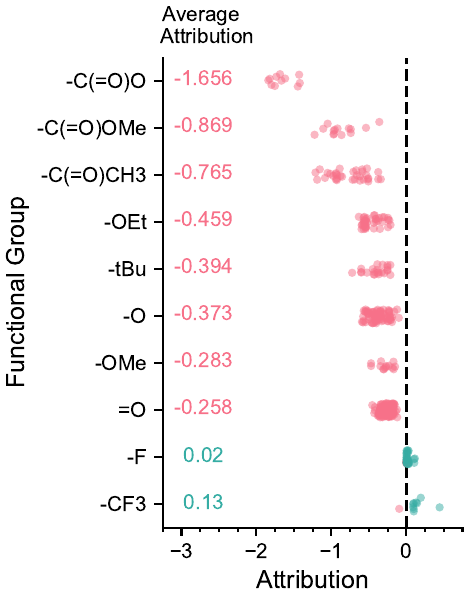}
     \caption{\justifying Swarm of functional group attribution to the predicted vapor pressure. 
     Most functional groups were attributed with lower vapor pressure, except for fluorine-containing groups (-F or -CF3).} 
     \label{fig:fg_attribution}
\end{figure}
\FloatBarrier

\subsection{\label{sec:Virtual Screening}Virtual Screening}
The GPR and GNN ensemble models were developed to guide the generation of new molecules and predict their vapor pressure.   
The interpretation of these models informed the inclusion of specific functional groups and the exclusion of fluorine in the generated molecules. 
For example, Figure~\ref{fig:fg_attribution} indicated that functional groups such as carboxylic acid, methyl ester, and methyl ketone are associated with reduced vapor pressure, and these groups were therefore included in molecule generation.

The generated molecules were created through a probabilistic growth procedure.
Seed structures were selected with equal probability between linear alkyl chains and cyclic structures. 
Molecules were then “grown” by replacing a randomly chosen hydrogen with either a cyclic structure (5\% probability), functional group (0 or 2\% probability) or an alkyl chain.
Growth stopped when the molecule either reached a molecular weight beyond 600 amu or 200 atoms. 
The number of carbon atoms in the additional alkyl chain was drawn from a Gaussian distribution ($\mu$ = 7, $\sigma$ = 3), bounded between 3 and 12 carbon atoms. 
The cyclic structure could have been as a cyclohexane, cyclopentane, benzene, toluene, and naphthalene.

The vapor pressure of the generated molecules was predicted using both the GPR and GNN ensemble models and benchmarked against the MAC molecules.
The GNN ensemble, trained to predict vapor pressure at 387~K, served as the initial filter.
At 387 K, MAC lubricants have vapor pressures near $10^{-5}$ Pa, so generated molecules with GNN-predicted vapor pressures higher than $10^{-5}$ Pa were removed from the candidate pool. 

The remaining molecules were then evaluated using the GPR ensemble, which can predict vapor pressure across a range of temperatures.
However, MD simulations at multiple temperatures are computationally expensive and were only performed at 300 K, a recommended temperature for reporting vapor pressure experimentally~\cite{ASTM-International2025-oz}.
At 300 K, MAC lubricants have a vapor pressure on the order of $5\times10^{-9}$ Pa~\cite{UnknownUnknown-lo, Jones2004-zw, Venier2003-zy, Clifford-Venier-Edward-W-Casserly-William-R-Jones-Jr-Mario-Marchetti-Mark-J-Jansen-Roamer-E-Predmore2002-bw,Nguyen2001-no,UnknownUnknown-lj}
, so molecules with GPR-predicted vapor pressures above $5\times10^{-9}$ Pa were removed from the candidate pool. 
From this process, more than 4500 molecules were identified with low vapor pressure for future studies. 

To explore structural diversity among the generated molecular candidates, the molecules were projected onto a low-dimensional chemical space and grouped according to structural similarity.
This process involved encoding molecular structures as Morgan fingerprints.
Pairwise structural dissimilarities were calculated using the Rogers-Tanimoto distance calculated from Morgan fingerprints~\cite{Bajusz2015-sa, Rogers1960-qh}.
The dissimilarities were then used as input for t-distributed Stochastic Neighbor Embedding (t-SNE)~\cite{Orlov2025-cf}.
The two-dimensional t-SNE projection positioned structurally similar molecules close together, allowing visualization of chemical diversity and structure–property relationships. 

Three distinct clusters within the projected space were identified using the Density Based Spatial Clustering of Applications with Noise (DBSCAN) algorithm~\cite{Ester1996-zp,Schubert2017-jm}.
Figure~\ref{fig:tSNE_withCentral} shows the t-SNE projection of low vapor pressure candidate molecular structures, clustered and distinguished by marker shape and color. 
Clusters 0, 1, and 2 contain 2754, 850, and 1345 molecules, respectively, which provides a large pool of candidate low vapor pressure molecular structures.
The median vapor pressures of the three clusters are $2.3\times10^{-9}$ Pa, $9.3\times10^{-10}$ Pa,  and $9.7\times10^{-10}$ Pa for cluster 0, 1 and 2, respectively. The standard deviation of vapor pressure in all clusters was less than 0.25 in log10 space. 
From the clusters, a representative molecule for each cluster was determined by calculating pairwise Rogers-Tanimoto similarity among all cluster members. 
The molecule with the highest mean similarity to all others was identified as the central or representative structure for each cluster.

The three representative molecules, R0, R1, and R2, are shown along with their chemical formulas in Figure~\ref{fig:CandidateMolecules}.
R0 represents a set of molecules that largely contain linear hydrocarbons with some H/C functional groups or double bonds. 
The R1 molecule represents a set of longer chain molecules that may or may not contain cyclic structures. 
Finally, the R2 molecule represents a set of highly branched molecules, which may or may not contain a cyclic structure as well. 
Structurally, the low vapor pressure 1,3-bis(2-octyldodecyl) cyclopentane and 1,3,4-tri-(2-octyldodecyl) cyclopentane are similar to the R1 and R2 molecules. 
The R0 is similar to waxy molecules, such as paraffin waxes~\cite{Kumar2016-ea}.
Although these are representative molecules, each cluster contains a distribution of chemically similar structures and properties. 
This analysis suggests chemistries that may have low vapor pressure and thus be viable space lubricants. 

\begin{figure}[!htbp]
     \centering
     \begin{subfigure}[b]{3.125in}
         \centering
         \includegraphics[width=\textwidth]{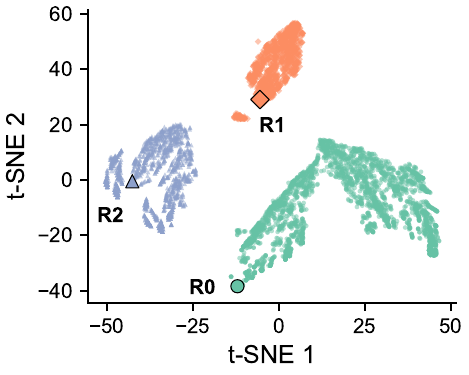}
         \caption{}
         \label{fig:tSNE_withCentral}
     \end{subfigure}
     \hfill
     \begin{subfigure}[b]{3.125in}
         \centering
         \includegraphics[width=\textwidth]{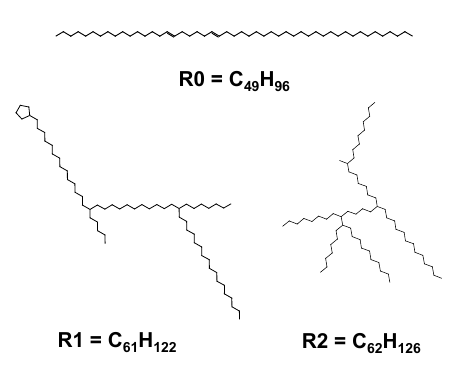}
         \caption{}
         \label{fig:CandidateMolecules}
     \end{subfigure}
     \hfill
     \caption{\justifying (a) t-SNE visualization of the chemical space and (b) representative low vapor pressure candidate molecular structures. In (a), the colors marker corresponds to clusters identified by DBSCAN algorithm. In (b), the representative molecular structures of each cluster are labeled with the molecular formula.} 
\end{figure}
\FloatBarrier
\section{\label{sec:conclusions}Conclusions}

This work demonstrated the integration of interpretable machine learning methods with molecular simulation data towards the discovery of space-compatible liquid-based lubricants. GPR and GNN ensemble models were used to predict vapor pressure across diverse chemical structures with high accuracy. The GPR models captured quantitative relationships between temperature, molecular density, and electronic features, while the GNN models provided structural attributions that highlighted the effects of functional groups such as carboxylic acids and esters in reducing vapor pressure.

The interpretability of these models provided valuable chemical insight into how molecular geometry, polarity, and composition influence volatility. Through virtual screening and clustering of generated molecules, a diverse set of candidates with predicted low vapor pressures was identified, extending the chemical design space beyond existing PAO, PFPE and MAC lubricants.
In future research, candidate molecules could be evaluated in terms of synthetic feasibility as well as other space lubricant-relevant properties such as viscosity.
More generally, the workflow developed here provides a generalizable and explainable approach for molecular property prediction and materials discovery in extreme environments. 

\FloatBarrier
\section*{Author Contributions}
DM performed the molecular dynamics simulations, curated the datasets, developed and implemented the machine learning models. DM and AM drafted the manuscript, and all authors revised and edited the text. AM supervised the project and acquired funding.

\section*{Conflicts of interest}
There are no conflicts to declare.

\section*{Acknowledgments}
The authors  would like to thank Andrew J. Clough and Peter P. Frantz for their motivation of this work and valuable discussion. 
This research was conducted using Pinnacles and MERCED clusters, which are maintained by the Cyberinfrastructure and Research Technologies (CIRT) at the University of California, Merced.
The MERCED cluster is centrally funded by the University of California, Merced and the Pinnacles cluster is supported by the NSF MRI grant (\#2019144) and the CENVAL\-ARC grant funded through the NSF CC* program (award \#2346744).
This work was supported by a NASA Space Technology Graduate Research Opportunity (NSTGRO) fellowship. 

\section*{Code Availability}
Code used in this study is available on the Github page: \url{https://github.com/Dmiliate2/ML4VP}. 

\section*{Data Availability}
The data used in this study is available on the Github page: \url{https://github.com/Dmiliate2/ML4VP}.

\balance

\bibliography{bibliography}
\bibliographystyle{bibstyle}

\end{document}


\makeatletter
\renewcommand{\@biblabel}[1]{\textsuperscript{S#1}} 
\renewcommand{\citenumfont}[1]{S#1}
\makeatother
\balance
\clearpage
\onecolumngrid 

\begin{center}
    \textbf{\huge Supplementary Information}\\[2ex]
    \LARGE\bfseries
    \MYTITLE
    
    \vspace{1em}
    
    \normalsize
    
    Daniel Miliate$^{1}$ and Ashlie Martini$^{1*}$
    
    \vspace{1em}

    \small
    $^{1}$University of California Merced, Merced, CA 95343, United States \\

\end{center}

\setcounter{equation}{0}
\setcounter{figure}{0}
\setcounter{table}{0}
\setcounter{section}{0}
\setcounter{page}{1} 

\renewcommand{\theequation}{S\arabic{equation}}
\renewcommand{\thefigure}{S\arabic{figure}}
\renewcommand{\thetable}{S\arabic{table}}
\renewcommand{\thesection}{S\arabic{section}}

\subsection{MAC Data}
\begin{figure*}[h]
    \centering
    \includegraphics[width=6.25in]{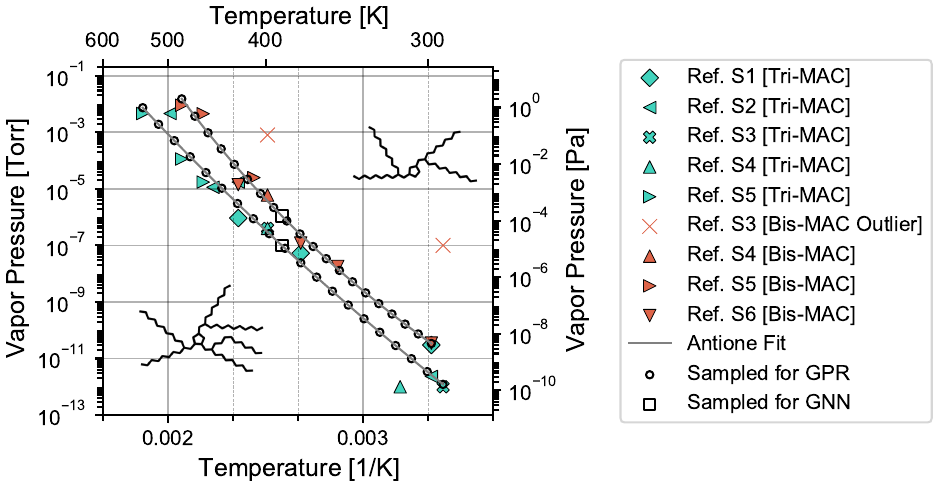}
    \caption{\justifying Vapor pressure vs temperature for 1,3,4-tri-(2-octyldodecyl) cyclopentane~\cite{UnknownUnknown-los, Jones2004-zws, Venier2003-zy, Clifford-Venier-Edward-W-Casserly-William-R-Jones-Jr-Mario-Marchetti-Mark-J-Jansen-Roamer-E-Predmore2002-bws,Nguyen2001-nos} and 1,3-bis(2-octyldodecyl) cyclopentane~\cite{Venier2003-zy, Clifford-Venier-Edward-W-Casserly-William-R-Jones-Jr-Mario-Marchetti-Mark-J-Jansen-Roamer-E-Predmore2002-bws,Nguyen2001-nos, UnknownUnknown-ljs}. 
    }
    \label{fig:MAC_DATA}
\end{figure*}

Data is limited for very low vapor pressure substances, but it is available for 1,3,4-tri-(2-octyldodecyl) cyclopentane~\cite{UnknownUnknown-los, Jones2004-zws, Venier2003-zy, Clifford-Venier-Edward-W-Casserly-William-R-Jones-Jr-Mario-Marchetti-Mark-J-Jansen-Roamer-E-Predmore2002-bws,Nguyen2001-nos} and bis(2-octyldodecyl) cyclopentane~\cite{Venier2003-zy, Clifford-Venier-Edward-W-Casserly-William-R-Jones-Jr-Mario-Marchetti-Mark-J-Jansen-Roamer-E-Predmore2002-bws,Nguyen2001-nos, UnknownUnknown-ljs} MAC lubricants. Data was extracted using WebPlotDigitizer~\cite{RohatgiUnknown-qls} when it was not tabulated. The colored markers in Figure~\ref{fig:MAC_DATA} correspond to reported data, where the marker shape corresponds to the source. Parameters for Antoine's equation were then calculated from the reported data, excluding outliers for 1,3-bis(2-octyldodecyl)~\cite{Venier2003-zy}. From the fit lines, vapor pressures and corresponding temperatures were sampled for training in GPR and GNN models.

\FloatBarrier

\subsection{Molecular Dynamics Simulation Protocol}

To generate high-throughput MD simulations, BOSS~\cite{Jorgensen2005-ql} was used to convert SMILES strings to structure files with OPLS-AA forcefields~\cite{Jorgensen1996-dh}.
OPLS-AA forcefields are widely accepted for non-reactive MD simulations.
The structure files were then used to build the simulation box with a maximum of 5000 atoms and 0.3 g/cc initial packing density, constructed by packmol~\cite{Martinez2009-mc}. 
The simulations were performed using the Large Atomic/Molecular Massively Parallel Simulator (LAMMPS) opensource software package~\cite{Thompson2022-md}. 
The MD simulation protocol was designed for relatively low computational costs, using both canonical (NVT) and isothermal–isobaric (NPT) ensembles and timestep of 1 fs.

The MD simulation began with energy minimization, setting the stopping tolerance for energy (unitless) at 1.0e-16, stopping tolerance for force at 1.06e-6 (kcal/mol)/Å, a maximum of 100,000 iterations, and a maximum of 500,000 evaluations. 
The simulation box was held at constant volume and constant temperature 1000 K with NVT for 50 ps. 
Next, the box was held at 1 atm constant pressure and constant temperature with NPT for 150 ps, ensuring density reached steady state. 
The simulation temperature of this step was chosen to match the temperature used in calculating vapor pressure. 
Finally, at the same pressure and temperature the NPT simulation trajectory was sampled every 0.5 ps for 50 ps to calculate dynamic features. 

\begin{figure*}[h]
     \centering
     \begin{subfigure}[b]{3.125in}
         \centering
         \includegraphics[width=\textwidth]{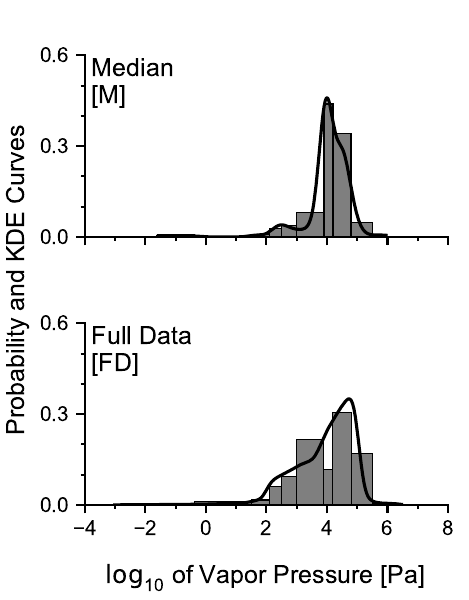}
         \caption{}
         \label{fig:PartDist_a}
     \end{subfigure}
     \hfill
     \begin{subfigure}[b]{3.125in}
         \centering
         \includegraphics[width=\textwidth]{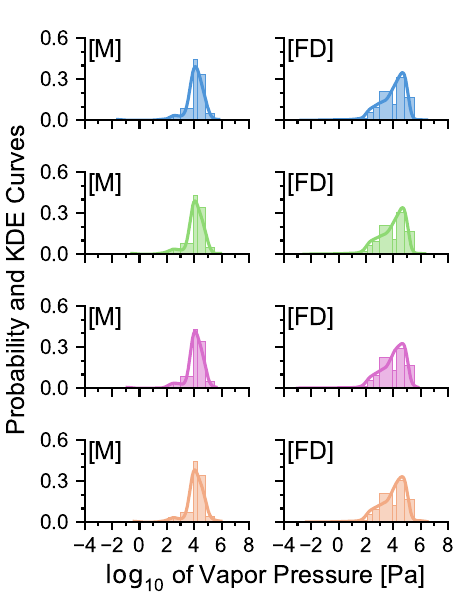}
         \caption{}
         \label{fig:PartDist_b}
     \end{subfigure}
     \hfill
     \caption{\justifying Overview probability histograms and kernel density estimate curves of the dataset and partitions for the median-of-molecular-structure [M] and full dataset [FD] of vapor pressure. In (a), the top and bottom histograms show the probability for the median and full dataset, respectively. In (b), each row represents the split partitions to be used for train \& validation sets of the outer CV and test set for final evaluation. The first and second columns represent the median-of-molecular-structure and full dataset, respectively.}
     \label{fig:Partition_Distribution}
\end{figure*}
\FloatBarrier

\begin{figure*}[h]
    \centering
    \includegraphics[width=6.25in]{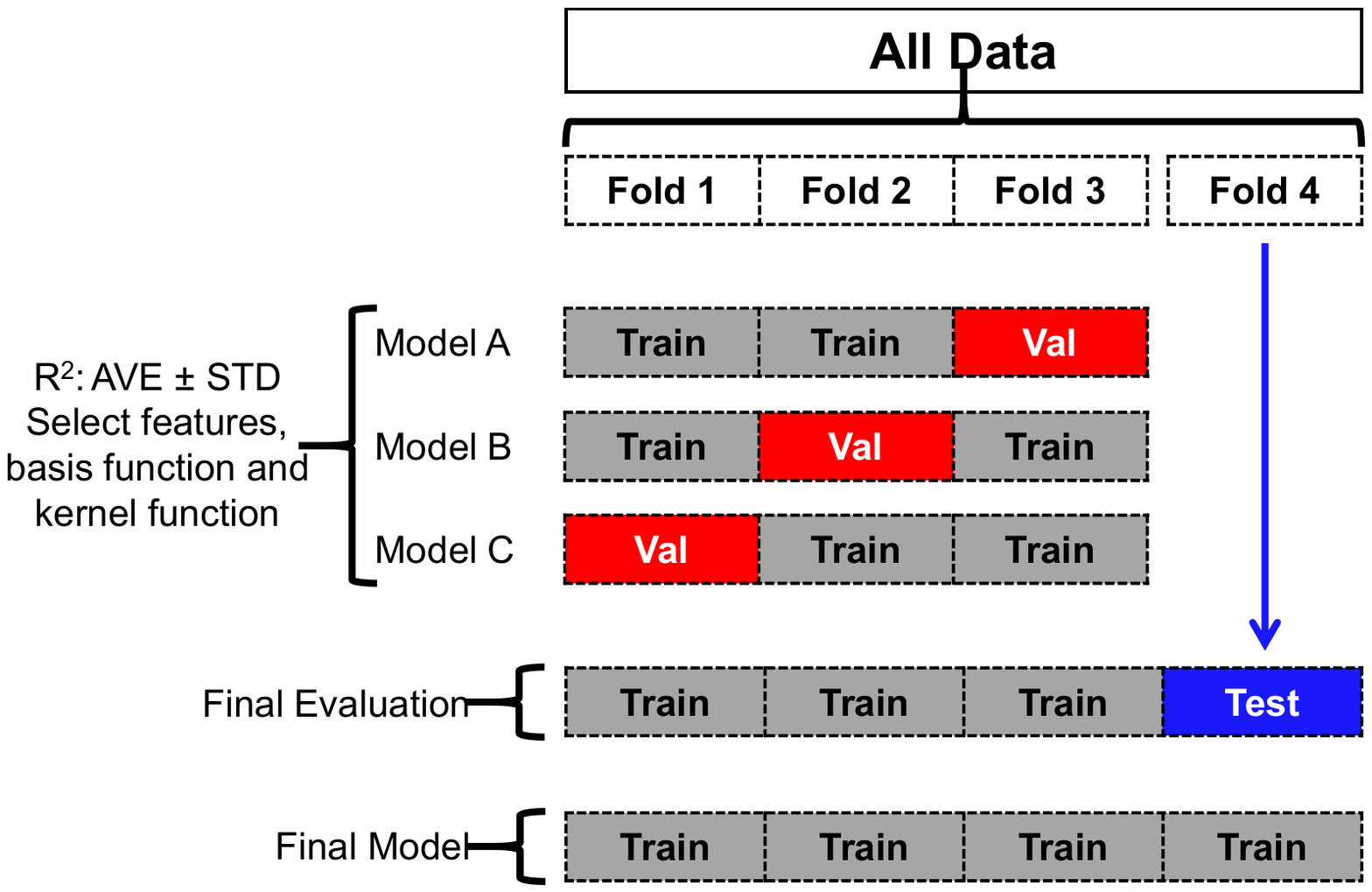}
    \caption{\justifying Data splitting workflow for nested CV model, final evaluation, and final model. During cross-validation, the best feature subsets, then basis and kernel functions are detemined before final evaluation.}
    \label{fig:Partition_Dataflow}
\end{figure*}
\FloatBarrier

\bibliography{SNotes}
\bibliographystyle{bibstyle}